\documentclass[final]{cvpr}

\usepackage{times}
\usepackage{epsfig}
\usepackage{graphicx}
\usepackage{amsfonts}
\usepackage{amsmath}
\usepackage{amssymb}
\usepackage{bbm}
\usepackage{multirow}
\usepackage{makecell}
\newcommand{\ignore}[1]{}
\usepackage[pagebackref=true,breaklinks=true,colorlinks,bookmarks=false]{hyperref}

\def\cvprPaperID{6901} 
\def\confYear{CVPR 2021}

\begin{document}

\title{Transformer Tracking}

\author{Xin Chen$^{1}$ \thanks{Equal contribution}, Bin Yan$^{1}$  \footnotemark[1], Jiawen Zhu$^{1}$, Dong Wang$^{1}$ \thanks{Corresponding author: Dr. Dong Wang, wdice@dlut.edu.cn}, Xiaoyun Yang$^{3}$ and Huchuan Lu$^{1,2}$\\
$^{1}$School of Information and Communication Engineering, Dalian University of Technology, China\\
$^{2}$Peng Cheng Laboratory $^{3}$Remark AI\\

{\tt\small \{chenxin3131, yan\_bin, jiawen\}@mail.dlut.edu.cn}\\ {\tt\small wdice@dlut.edu.cn, xyang@remarkholdings.com, lhchuan@dlut.edu.cn}
}

\maketitle

\begin{abstract}
   Correlation acts as a critical role in the tracking field, especially in recent popular Siamese-based trackers. 
   The correlation operation is a simple fusion manner to consider the similarity between the template and the search region. 
   However, the correlation operation itself is a local linear matching process, leading to lose semantic information and fall into local optimum easily, which may be the bottleneck of designing high-accuracy tracking algorithms. 
   Is there any better feature fusion method than correlation? 
   To address this issue, inspired by Transformer, this work presents a novel attention-based feature fusion network, 
   which effectively combines the template and search region features solely using attention. 
   Specifically, the proposed method includes an ego-context augment module based on self-attention and a cross-feature 
   augment module based on cross-attention. 
   Finally, we present a Transformer tracking (named TransT) method based on the Siamese-like feature 
   extraction backbone, the designed attention-based fusion mechanism, and the classification and regression head. 
   Experiments show that our TransT achieves very promising results on six challenging datasets, 
   especially on large-scale LaSOT, TrackingNet, and GOT-10k benchmarks. 
   Our tracker runs at approximatively 50 $fps$ on GPU. Code and models are available at \href{https://github.com/chenxin-dlut/TransT}{https://github.com/chenxin-dlut/TransT}.
\end{abstract}

\thispagestyle{empty}

\section{Introduction}

\begin{figure}[!t]
\begin{center}
\resizebox{\linewidth}{!}{
\begin{tabular}{p{2cm}<{\centering}@{}p{4.2cm}<{\centering}@{}p{2cm}<{\centering}}
\scriptsize{Self-attention} & \scriptsize{Screen Shots} & \scriptsize{Cross-attention}\\
\end{tabular}}
\includegraphics[width=1\linewidth]{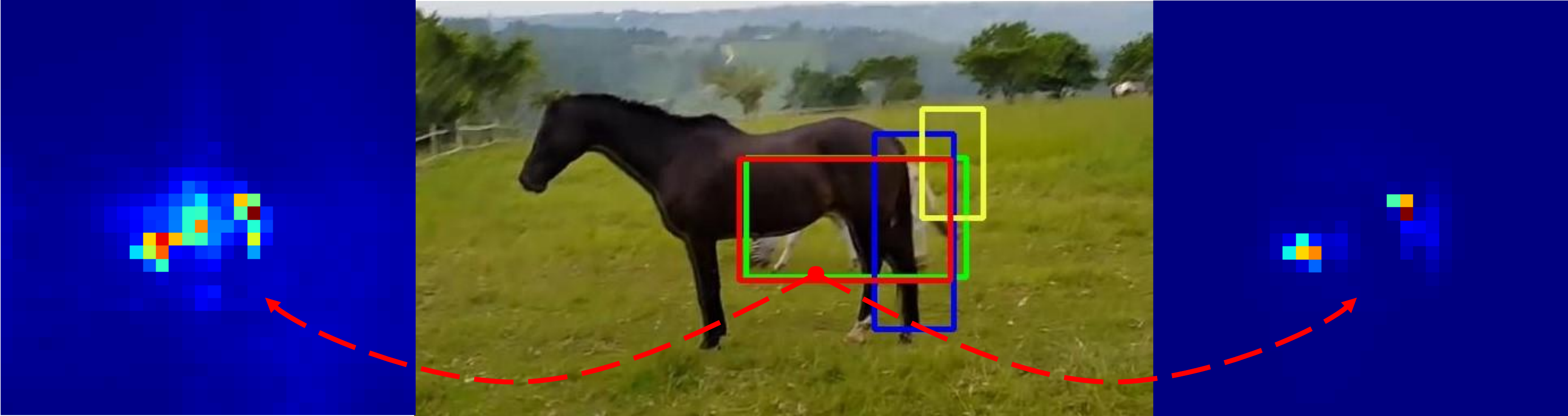}
\end{center}
\vspace{-0.75cm}
\begin{center}
\includegraphics[width=1\linewidth]{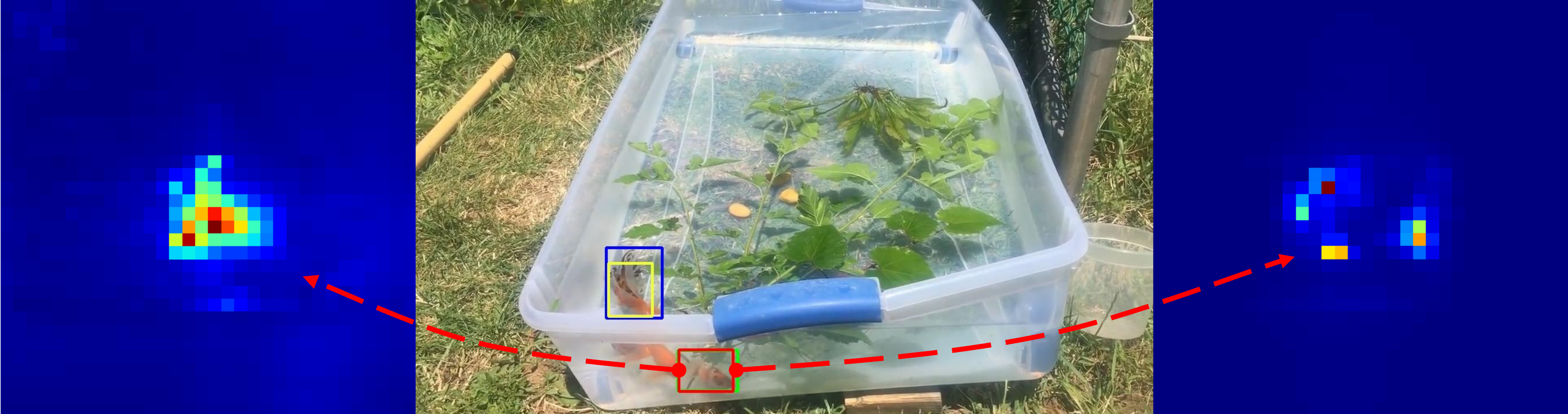}
\end{center}
\vspace{-0.75cm}
\begin{center}
\includegraphics[width=1\linewidth]{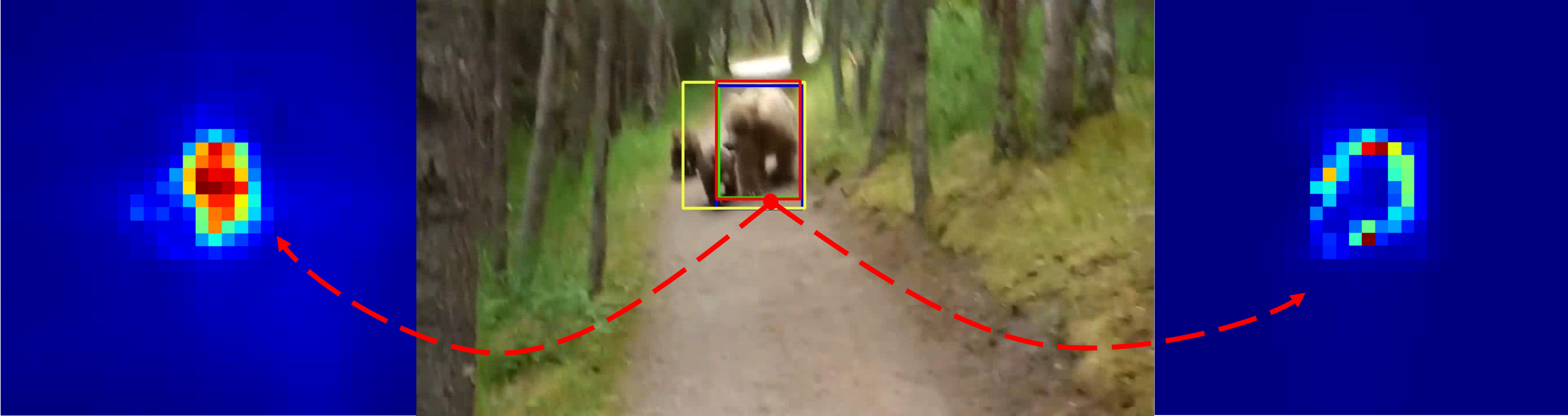}
\end{center}
\vspace{-0.8cm}
\begin{center}
\includegraphics[width=1\linewidth]{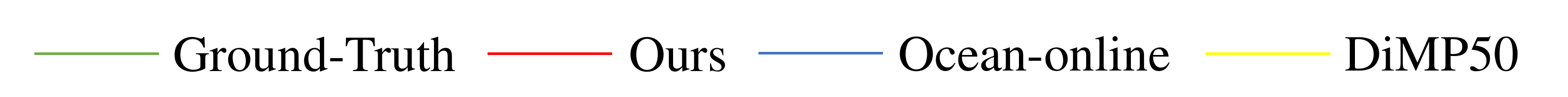}
\end{center}
\vspace{-2mm}
\caption{Tracking results of TransT and two state-of-the-art trackers. Our tracker is more robust 
and accurate in handling various challenges, such as occlusion, similar object interference, motion blur.}
\label{fig:visual}
\end{figure}

Visual object tracking is a fundamental task in computer vision, which aims to predict 
the position and shape of a given target in each video frame. 
It has a wide range of applications in robot vision, video surveillance, unmanned driving,  
and other fields. 
The main challenges of tracking are large occlusion, severe deformation, interference from 
similar objects, to name a few. 
Many efforts have been done in recent years~\cite{DVT-Review,DVT-review-new}, however, designing 
a high-accuracy and real-time tracker is still a challenging task. 

For most of the popular trackers (such as SiamFC~\cite{SiameseFC}, SiamRPN~\cite{SiameseRPN}, and ATOM~\cite{ATOM}), correlation 
plays a critical role in integrating the template or target information into the regions of interest (ROI). 
However, the correlation operation itself is a linear matching process and leads to semantic information loss, 
which limits the tracker to capture the complicated non-linear interaction between the template and ROIs. 
Thus, previous models have to improve the non-linear representation ability by introducing fashion
structures~\cite{SiamRPNplusplus,SiamFC++,Ocean}, using additional modules~\cite{MLT,DSA,CGACD}, 
designing effective online updaters~\cite{DiMP,UpdateNet,PrDiMP}, to name a few.  
This naturally introduces an interesting question: is there any better feature fusion method than correlation?

In this work, inspired by the core idea of Transformer~\cite{2017Attention}, we address the aforementioned 
problem by designing an attention-based feature fusion network and propose a novel Transformer tracking 
algorithm (named TransT).  
The proposed feature fusion network consists of an ego-context augment module based on self-attention 
and a cross-feature augment module based on cross-attention. 
This fusion mechanism effectively integrates the template and ROI features, producing more semantic 
feature maps than correlation. 
Figure~\ref{fig:visual} provides some representative visual results, illustrating that our TransT method 
produces insightful attention maps regarding the target and performs better than other competing trackers. 
Our main contributions are summarized as follows.

\vspace{-1mm}
\begin{itemize}
    \setlength{\itemsep}{0pt}
    \setlength{\parsep}{0pt}
    \setlength{\parskip}{0pt}
	\item We propose a novel Transformer tracking framework, consisting of feature extraction, Transformer-like fusion, 
	and head prediction modules. The Transformer-like fusion combines the template and search region features 
	solely using attention, without correlation.  
	\item We develop our feature fusion network based on an ego-context augment module with self-attention as 
	well as a cross-feature augment module with cross-attention. Compared with correlation-based feature fusion, 
	our attention-based method adaptively focuses on useful information, such as edges and similar targets, and establishes associations between distant features, 
	to make the tracker obtain better classification and regression results. 
    \item Numerous experimental results on many benchmarks show that the proposed tracker performs significantly better than 
    the state-of-the-art algorithms, especially on large-scale LaSOT, TrackingNet, GOT-10k datasets. Besides, our tracker runs at about 50 $fps$ in GPU, which meets the real-time requirement. 
\end{itemize} 

\section{Related Work}
{\noindent \textbf{Visual Object Tracking.}}
In recent years, Siamese-based methods have been more popular in the tracking field~\cite{SINT,SiameseFC,SiameseRPN,SiamMask,SiamRPNplusplus,SiamFC++,Ocean}. 
SiamFC~\cite{SiameseFC}, the pioneering work, combines naive feature correlation with Siamese framework. 
After that, SiamRPN~\cite{SiameseRPN} combines the Siamese network with RPN~\cite{FasterRCNN} and conducts feature fusion using 
depthwise correlation, to obtain more precise tracking results. 
Some further improvements have been made, such as adding additional branches~\cite{SiamMask,Alpha-Refine}, using deeper architectures~\cite{SiamRPNplusplus}, 
exploiting anchor-free architectures~\cite{SiamFC++,Ocean}, and so on. 
These mainstream tracking architectures can be divided into two parts: a backbone network to extract image features, 
followed by a correlation-based network to compute the similarity between the template and the search region. 
Moreover, some popular online trackers (e.g., ECO~\cite{ECO}, ATOM~\cite{ATOM}, and DiMP~\cite{DiMP}) also heavily rely on the correlation 
operation. 
However, two issues have been overlooked. First, the correlation-based network does not make full use of global context, so it is easy to fall into the local optimum. Second, through correlation, the semantic information has been lost to some degree, which 
may lead to an imprecise prediction regarding the target's boundaries. 
Therefore, in this work, we design a variant structure of Transformer based on attention to replace the correlation-based network 
for conducting feature fusion. 
\begin{figure*}[!t]
\begin{center}
\includegraphics[width=1\linewidth]{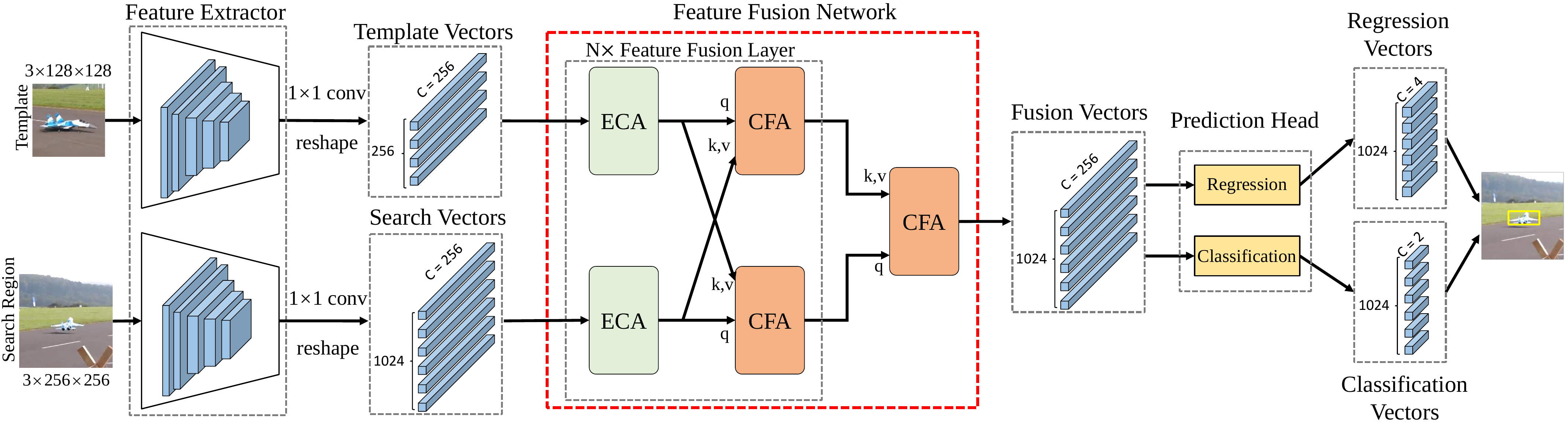}
\end{center}
   \caption{Architecture of our Transformer tracking framework. This framework contains three fundamental components: 
   feature extraction backbone, feature fusion network, and prediction head. The proposed attention-based feature fusion 
   network is naturally applied on the Siamese-based feature extraction backbone.}
\label{fig:TransT}
\end{figure*}

{\noindent \textbf{Transformer and Attention.}} 
Transformer~\cite{2017Attention} was first introduced by Vaswani \emph{et al.} and applied in machine translation. 
Briefly, Transformer is an architecture for transforming one sequence into another one with the help of attention-based 
encoders and decoders. 
The attention mechanism looks at an input sequence and decides at each step which other parts of the sequence are important, 
and therefore facilitates capturing the global information from the input sequence. 
Transformer has replaced recurrent neural networks in many sequential tasks (natural language processing~\cite{BERT}, 
speech processing~\cite{LibriSpeech,ASR}, and computer vision~\cite{ImgTrans}), and gradually extended to handle 
non-sequential problems~\cite{ViT,DETR}.
In~\cite{DETR}, Carion \emph{et al.} considers object detection as 
a set prediction problem and adopts the encoder-decoder architecture in~\cite{2017Attention} as the detection head. 
Experiments on COCO~\cite{COCO} demonstrate that the DETR approach achieves comparable results to an optimized Faster R-CNN 
baseline~\cite{FasterRCNN}.
Motivated by the success of DETR 
as well as the close relationship between detection and tracking (like RPN~\cite{FasterRCNN} and SiamRPN~\cite{SiameseRPN}), we attempt to introduce Transformer into the tracking field. 
Different from DETR, we do not directly  
follow the encoder-decoder architecture in the original Transformer 
as it is not very matched with the tracking task. 
We adopt the core idea of Transformer and exploit the attention mechanism to design the ego-context augment (ECA) and 
cross-feature augment (CFA) modules. 
The integration of ECA and CFA focuses on feature fusion between template and search region, rather than extracting  
information from only one image in~\cite{DETR}. 
This design philosophy is more suitable for visual object tracking.

Several efforts have been made to introduce the attention mechanism in the tracking field. 
ACF~\cite{ACF} learns an attention network to do switching among different correlation filters. 
MLT~\cite{MLT} adopts channel-wise attention to provide the matching network with target-specific information. 
These two works merely borrow the concept of attention to conduct model or feature selection. 
For improving the tracking performance, different attention layers (such as channel-wise attention~\cite{RASNet,SA-Siam}, 
spatial-temporal attention~\cite{CFSTA}, and residual attention~\cite{RASNet}) are utilized to enhance the template 
information within the correlation matching framework.
SiamAttn~\cite{DSA} explores both self-attention and cross branch attention to improve the discriminative ability of 
target features before applying depth-wise cross correlation. 
CGACD~\cite{CGACD} learns attention from the correlation result of the template and search region, and then 
adopts the learned attention to enhance the search region features for further classification and regression. 
These works have improved the tracking accuracy with the attention mechanism, but they still highly rely on the 
correlation operation in fusing the template and search region features. 
In this work, we exploit the core idea of Transformer and design a new attention-based network to directly fuse template 
and search region features without using any correlation operation. 
\section{Transformer Tracking}
\label{sec:TransT}
This section presents the proposed Transformer Tracking method, named TransT. 
As shown in Figure~\ref{fig:TransT}, our TransT is very concise, consisting of three components: 
backbone network, feature fusion network and prediction head. 
The backbone network separately extracts the features of the template and the search region. 
Then, the features are enhanced and fused by the proposed feature fusion network. 
Finally, the prediction head performs the binary classification and bounding box regression on 
the enhanced features to generate the tracking results\footnote{The tracking results are also post-processed 
by the window penalty, which will be introduced in Section~\ref{sec-exp}.}.
We introduce the details of each component of our TransT, introduce the two important modules 
in the feature fusion network, and then provide some illustrations and discussions.

\subsection{Overall Architecture}
{\noindent \textbf{Feature Extraction. }}
Like Siamese-based trackers~\cite{SiameseFC,SiameseRPN}, the proposed TransT method takes a pair of image patches 
(i.e., the template image patch $z \in {\mathbb{R}}^{3 \times {H_{z0}} \times {W_{z0}}}$ and the search region 
image patch $x \in {\mathbb{R}}^{3 \times {H_{x0}} \times {W_{x0}}}$) as the inputs of the backbone network. 
The template patch is expanded by twice the side length from the center of the target in the first frame of 
a video sequence, which includes the appearance information of the target and its local surrounding scene. 
The search region patch is expanded four times the side length from the center coordinate of the target 
in the previous frame, and the search region typically covers the possible moving range of the target. 
Search region and template are reshaped to squares, then be processed by the backbone. 
We use a modified version of ResNet50~\cite{ResNet} for feature extraction. 
Specifically, we remove the last stage of ResNet50 and take the outputs of the fourth stage as final 
outputs. We also change the convolution stride of the down-sampling unit of the fourth stage from 
2 to 1, to obtain a larger feature resolution. 
Besides, we modify the $3 \times 3$ convolution in the fourth stage to dilation convolution with 
stride of 2 to increase the receptive field. 
The backbone processes the search region and the template to obtain their features maps 
${f_z} \in {\mathbb{R}}^{C \times {H_z} \times {W_z}}$ and ${f_x} \in {\mathbb{R}}^{C \times {H_x} \times {W_x}}$.
$H_z,W_z = {\frac{H_{z0}}{8}},{\frac{W_{z0}}{8}}$, $H_x,W_x = {\frac{H_{x0}}{8}},{\frac{W_{x0}}{8}}$ and $C = 1024$.

{\noindent \textbf{Feature Fusion Network. }}
We design a feature fusion network to effectively fuse the features $f_z$ and $f_x$. First, a $1 \times 1$ convolution reduces the channel dimension of $f_z$ and $f_x$, obtaining two lower dimension feature maps,  $f_{z0} \in \mathbb{R}^{d \times H_z \times W_z}$ and $f_{x0} \in \mathbb{R}^{d \times H_x \times W_x}$. We employ $d=256$ in our implementation. Since the attention-based feature fusion network takes a set of feature vectors as input, we flatten $f_{z0}$ and $f_{x0}$ in spatial dimension, obtaining $f_{z1} \in \mathbb{R}^{d \times H_{z}W_{z}}$ and $f_{x1} \in \mathbb{R}^{d \times H_{x}W_{x}}$.  Both $f_{z1}$ and $f_{x1}$ can be regarded as a set of feature vectors of length $d$.  
As shown in Figure~\ref{fig:TransT}, the feature fusion network takes $f_{z1}$ and $f_{x1}$ as the inputs to 
the template branch and the search region branch respectively.
First, two ego-context augment (ECA) modules focus on the useful 
semantic context adaptively by multi-head self-attention, to enhance the feature representation.
Then, two cross-feature augment (CFA) modules receive the feature maps of their own and the other branch 
at the same time and fuse these two feature maps through multi-head cross-attention. 
In this way, two ECAs and two CFAs form a fusion layer, as shown in the dotted box in Figure~\ref{fig:TransT}. 
The fusion layer repeats $N$ times, followed by an additional CFA to fuse the feature map of two branches, 
decoding a feature map $f \in \mathbb{R}^{d \times H_{x}W_{x}}$ (we employ $N=4$ in this work). 
The details of ECA and CFA modules are introduced in Section~\ref{section:CFAECA}.

{\noindent \textbf{Prediction Head Network.}} 
The prediction head consists of a classification branch and a regression branch, where each branch is a three-layer perceptron with hidden dimension $d$ and a $\rm{ReLU}$ activation function. 
For the feature map $f \in \mathbb{R}^{d \times H_{x}W_{x}}$ generated by the feature fusion network, the head makes predictions on 
each vector to get $H_{x}W_{x}$ foreground/background classification results, and $H_{x}W_{x}$ normalized coordinates with respect 
to the search region size. 
Our tracker directly predicts the normalized coordinates instead of adjusting the anchor points or anchor boxes, completely discarding the anchor points or anchor boxes based on prior knowledge, thereby making the tracking framework more concise.

\subsection{Ego-Context Augment and Cross-Feature Augment Modules}
\label{section:CFAECA}
{\noindent \textbf{Multi-head Attention. }}
Attention is the fundamental component in designing our feature fusion network. 
Given queries $\mathbf{Q}$, keys $\mathbf{K}$ and values $\mathbf{V}$, the attention 
function is the scale dot-product attention, defined in equation (\ref{eq-s-att}).
\begin{equation}
    {\rm{Attention}}(\mathbf{Q},\mathbf{K},\mathbf{V})
    = {\rm{softmax}}(\frac{\mathbf{Q}\mathbf{K}^\top}{\sqrt{d_k}})\mathbf{V}, 
\label{eq-s-att}
\end{equation}
where $d_k$ is the key dimensionality. 

As described in~\cite{2017Attention}, extending the attention mechanism (\ref{eq-s-att}) 
into multiple heads enable the mechanism to consider various attention distributions and make the model 
pay attention to different aspects of information. The mechanism of multi-head attention 
is defined in equation (\ref{eq-m-att}). We refer the reader to the literature~\cite{2017Attention} 
for more detailed descriptions. 
\begin{equation}
\label{eq-m-att}
{\rm{MultiHead}}(\mathbf{Q},\mathbf{K},\mathbf{V}) = {\rm{Concat}}({\mathbf{H}_1},...,{\mathbf{H}_{{n_h}}}){\mathbf{W}^O}, 
\end{equation}
\begin{equation}
{\mathbf{H}_i} = {\rm{Attention}}(\mathbf{Q}\mathbf{W}_i^Q,\mathbf{K}\mathbf{W}_i^K,
\mathbf{V}\mathbf{W}_i^V),
\end{equation}
where $\mathbf{W}_i^Q \in \mathbb{R}^{d_m \times d_k}$, $\mathbf{W}_i^K \in \mathbb{R}^{d_m \times d_k}$, $\mathbf{W}_i^V \in \mathbb{R}^{d_m \times d_v}$, and $\mathbf{W}^O \in \mathbb{R}^{n_hd_v \times d_m}$ are parameter matrices. 
In this work, we employ $n_h=8$, $d_m=256$ and 
${{d_k} = {d_v} = {{{d_m}} \mathord{\left/
 {\vphantom {{{d_m}} {{n_h}}}} \right.
 \kern-\nulldelimiterspace} {{n_h}}} = 32}$
as default values.

\begin{figure}[t]
\begin{center}
\includegraphics[width=1\linewidth]{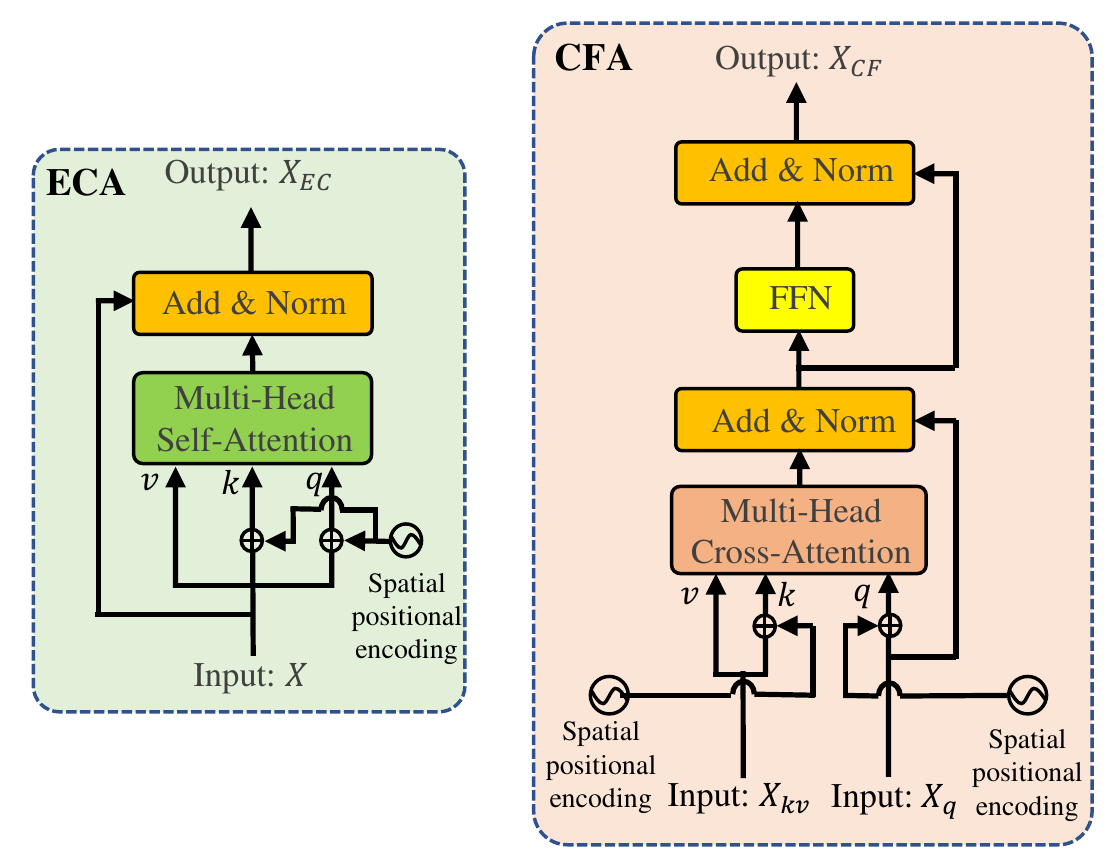}
\end{center}
   \caption{Left: ECA module. Right: CFA module. The ECA module is based on multi-head self-attention in a residual form. The CFA module is based on multi-head cross-attention and FFN in a residual form. The input $X_q$ receives the feature from the branch where CFA is located, and $X_{kv}$ receives the feature from the other branch. Spatial positional encodings are used to encode position information. ECA enhances the contextual information of the input and CFA adaptively fuses the features from two branches.}
\label{fig:ECACFA}
\end{figure}

{\noindent \textbf{Ego-Context Augment (ECA). }}
The structure of ECA is shown in the left of Figure~\ref{fig:ECACFA}. 
ECA adaptively integrates the information from different positions of the feature map, 
by using multi-head self-attention in the form of residual. 
As shown in equation (\ref{eq-s-att}), the attention mechanism has no ability to distinguish 
the position information of the input feature sequence. 
Thus, we introduce a spatial positional encoding process to the input $\mathbf{X} \in \mathbb{R}^{d \times N_x}$. 
Following \cite{DETR}, we use a sine function to generate spatial positional encoding. 
Finally, the mechanism of ECA can be summarized as 
\begin{equation}
\begin{split}
\label{equation:Multi-Head Attention}
{{\bf{X}}_{EC}} = {\bf{X}} + {\rm MultiHead}({\bf{X}} + {{\bf{P}}_x},{\bf{X}} + {{\bf{P}}_x},{\bf{X}})
\end{split}, 
\end{equation}
where $\mathbf{P}_x \in \mathbb{R}^{d \times N_x}$ is the spatial positional encodings and $\mathbf{X}_{EC} \in 
\mathbb{R}^{d \times N_x}$ is the output of ECA.

{\noindent \textbf{Cross-Feature Augment (CFA). }}
The structure of CFA is shown in the right of Figure~\ref{fig:ECACFA} . 
CFA fuses the feature vectors from two inputs by using multi-head cross-attention in the form of residual. 
Similar to ECA, spatial positional encoding is also used in CFA. 
In addition, a FFN module is used to enhance the fitting ability of the model, which is a fully connected 
feed-forward network that consists of two linear transformation with a $\rm{ReLU}$ in between, that is,
\begin{equation}
{\rm FFN}\left( {\bf{x}} \right) = \max \left( {{\bf{0}},{\bf{x}}{{\bf{W}}_1} + 
{{\bf{b}}_1}} \right){{\bf{W}}_2} + {{\bf{b}}_2},
\end{equation}
the symbols $\mathbf{W}$ and $b$ stand for weight matrices and basis vectors, respectively. 
The subscripts denote different layers.

Thus, the mechanism of CFA can be summarized as 
\begin{equation}
\begin{array}{c}
{{\bf{X}}_{CF}} = {\widetilde{\bf{X}}_{CF}} + {\rm{FFN}}\left( {{{\widetilde{\bf{X}}}_{CF}}} \right), \\
{\widetilde{\bf{X}}_{CF}} = {{\bf{X}}_q} + {\rm{MultiHead}}\left( {{{\bf{X}}_q} + {{\bf{P}}_q},{{\bf{X}}_{kv}} + {{\bf{P}}_{kv}},{{\bf{X}}_{kv}}} \right), 
\end{array}
\label{eq-cfa}
\end{equation}
where $\mathbf{X}_q \in \mathbb{R}^{d \times N_q}$ is the input of the branch 
where the module is applied, $\mathbf{P}_q \in \mathbb{R}^{d \times N_q}$ is the spatial 
positional encoding corresponding to $\mathbf{X}_q$. 
$\mathbf{X}_{kv} \in \mathbb{R}^{d \times N_{kv}}$ is the input from another branch, 
and $\mathbf{P}_{kv} \in \mathbb{R}^{d \times N_{kv}}$ is the spatial encoding for the coordinate of $\mathbf{X}_{kv}$. 
$\mathbf{X}_{CF} \in \mathbb{R}^{d \times N_q}$ is the output of CFA. 
According to equation (\ref{eq-cfa}), CFA calculates the attention map according to multiple scaled products between $\mathbf{X}_{kv}$ and $\mathbf{X}_q$, then reweighs $\mathbf{X}_{kv}$ according to the attention map, 
and adds it to $\mathbf{X}_q$ to enhance the representation ability of the feature map.

{\noindent \textbf{Differences with the original Transformer. }}
Our method draws on the core idea of Transformer, i.e., employing the attention mechanism. 
But we do not directly adopt the structure of the Transformer in DETR~\cite{DETR}. 
Instead, we design a new structure to make it more suitable for tracking framework. 
The cross-attention operation in our method plays a more important role than that in DETR, since 
the tracking task focuses on fusing the template and search region features. 
Experimental comparisons of the trackers with our method and the original Transformer 
are shown in Section~\ref{subsec-as}.

\begin{figure}[t]
\begin{center}
\includegraphics[width=1\linewidth]{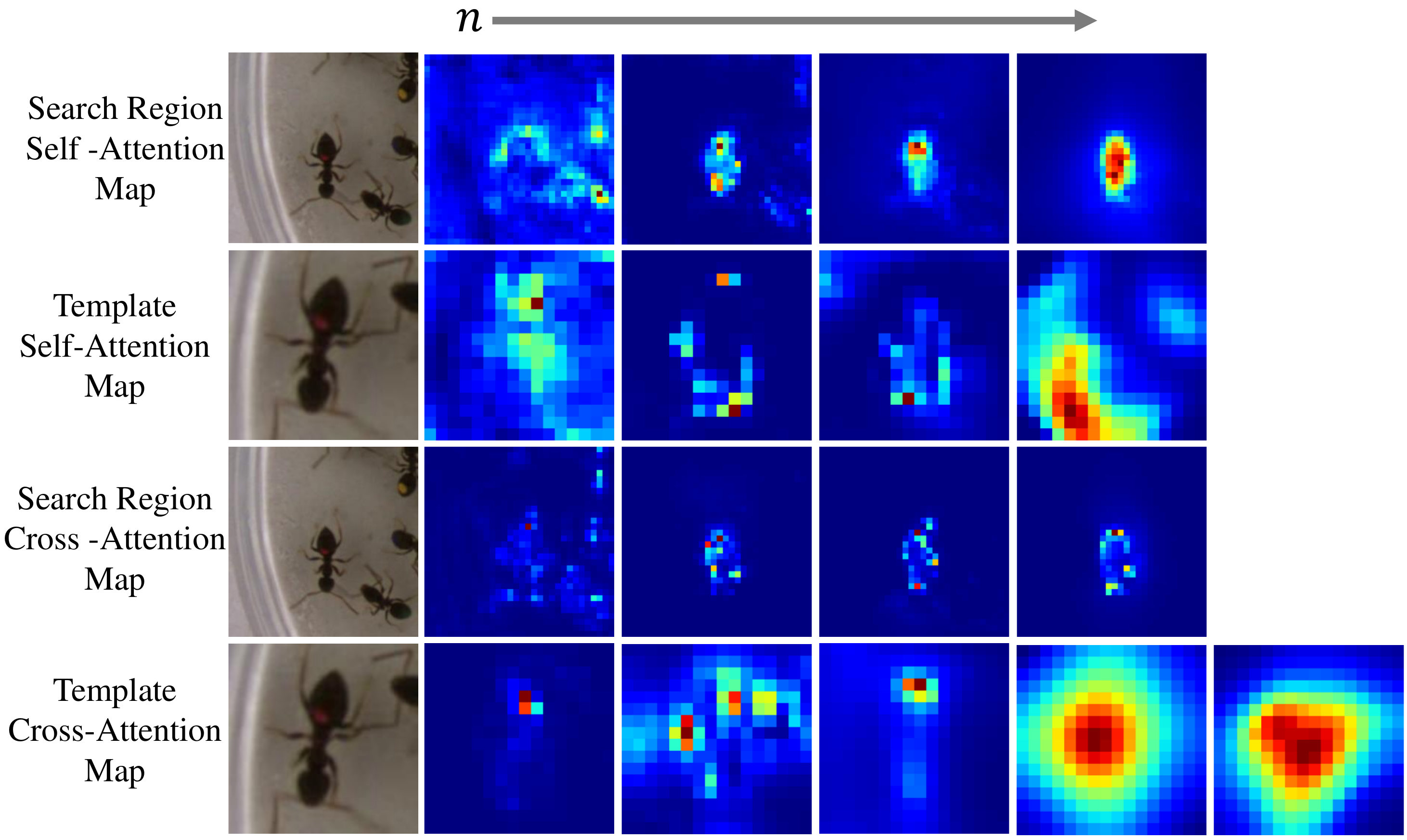}
\end{center}
   \caption{Visualization of the attention maps for a representative pair. From left to right, the feature fusion 
   layer goes deeper. From top to bottom, they are self-attention maps in the search region, self-attention maps 
   in the template, cross-attention maps in the search region, and cross-attention maps in the template, respectively.}
\label{fig:Attentionmap}
\end{figure}

{\noindent \textbf{What does attention want to see?}} 
To explore how the attention module works in our framework, we visualized the attention maps of all attention 
modules in a representative tracking clip, as shown in~Figure~\ref{fig:Attentionmap}, to see what the attention wants to see. 
We use the number $n$ ($1{\leq}n{\leq}4$) to represent the current number of the fusion layer. 
There are four layers in total, and the fusion layer goes deeper from left to right.  
The last single attention map is obtained from the last cross-attention, which is used for decoding. 

The first line shows self-attention maps of the search region. When $n=1$, there is no information from the template, the attention module tries to see all objects that are different from the environment. 
The same thing happens in the second line, i.e., self-attention map of template. 
Interestingly, attention focuses more on key information, such as the red dot on the tail of the ant. 
The third and fourth lines are cross-attention maps applied to the search region and template respectively. 
At this point, attention modules receive features from both template and search region. 
To locate the target under the interference of similar targets, attention modules tend to pay attention 
to the important information, i.e., the colored points on the tail of ants.
When $n=2$, at this point, the inputs of every attention module have fused the search region and template information. 
The focus of the search region self-attention map on similar distractors has been reduced, the model appears to have 
recognized the target. The cross-attention map to the search region seems quite sure of its estimation. 
For the template, attention modules begin to focus on boundary information. 

As the fusion layers go deeper, the search region self-attention map tends to strengthen the 
location of the target, while the cross-attention map to the search region focuses on the boundary of the identified target. 
In this way, the template feature becomes an information bank that contains a large amount of the target's boundary 
information, while the search region feature still keeps its spatial information. We notice that the last few attention 
maps for the template no longer follow the initial spatial position, but a puzzling distribution. 
Perhaps this is because, after the target has been identified, the features of the template branch no longer need to 
keep the information of the template itself, but store a lot of the target's boundary information, becoming a feature 
library serving for regression. Through the visualization of the attention maps, we can see that the attention 
modules automatically look for global useful information, thereby making the tracker achieve good results. 

\begin{table*}[!t]
\caption{State-of-the-art comparison on TrackingNet, LaSOT, and GOT-10k. The best two results 
are shown in \textbf{\textcolor{red}{red}} and \textbf{\textcolor{blue}{blue}} fonts.}
\label{tab-sota1}
\vspace{-3mm}
\begin{center}
\resizebox{\linewidth}{!}{
\begin{tabular}{|c|c|ccc|ccc|ccc|}
\hline
\multirow{2}{*}{Method} & \multirow{2}{*}{Source} &\multicolumn{3}{|c|}{LaSOT~\cite{LaSOT}}	
&\multicolumn{3}{|c|}{TrackingNet~\cite{trackingnet}}	&\multicolumn{3}{|c|}{GOT-10k~\cite{GOT10K}}\\
\cline{3-11}
& &AUC	&P$_{Norm}$	&P	&AUC	&P$_{Norm}$	&P	&AO	&SR$_{0.5}$	&SR$_{0.75}$\\
\hline
TransT	&Ours	&\textcolor{red}{\textbf{64.9}}	&\textcolor{red}{\textbf{73.8}}	&\textcolor{red}{\textbf{69.0}}	
&\textcolor{red}{\textbf{81.4}}	&\textcolor{red}{\textbf{86.7}}	&\textcolor{red}{\textbf{80.3}}	
&\textcolor{red}{\textbf{72.3}}	&\textcolor{red}{\textbf{82.4}}	&\textcolor{red}{\textbf{68.2}}\\
TransT-GOT	&Ours	&-	&-	&-	
&-	&-	&-	
&\textcolor{blue}{\textbf{67.1}}	
&\textcolor{blue}{\textbf{76.8}}	
&\textcolor{blue}{\textbf{60.9}}\\
SiamR-CNN~\cite{SiamRCNN}   &CVPR2020	&\textcolor{blue}{\textbf{64.8}}	
&\textcolor{blue}{\textbf{72.2}}	&-	&\textcolor{blue}{\textbf{81.2}}	
&\textcolor{blue}{\textbf{85.4}}	
&\textcolor{blue}{\textbf{80.0}}		&64.9	&72.8	&59.7\\
Ocean~\cite{Ocean}	&ECCV2020	&56.0	&65.1	&56.6	&-	&-	&-	&61.1	&72.1	&47.3\\
KYS~\cite{KYS}	    &ECCV2020	&55.4	&63.3	&-	&74.0	&80.0	&68.8	&63.6	&75.1	&51.5\\
DCFST~\cite{DCFST}	&ECCV2020	&-	&-	&-	&75.2	&80.9	&70.0	&63.8	&75.3	&49.8\\
SiamFC++~\cite{SiamFC++}	&AAAI2020	&54.4	&62.3	&54.7	&75.4	&80.0	&70.5	&59.5	&69.5	&47.9\\
PrDiMP~\cite{PrDiMP}&CVPR2020	&59.8	&68.8	&60.8	&75.8	&81.6	&70.4	&63.4	&73.8	&54.3\\
CGACD~\cite{CGACD}	&CVPR2020	&51.8	&62.6	&-	&71.1	&80.0	&69.3	&-	&-	&-\\
SiamAttn~\cite{DSA}	&CVPR2020	&56.0	&64.8	&-	&75.2	&81.7	&-	&-	&-	&-\\
MAML~\cite{MAML}	&CVPR2020	&52.3	&-	&-	&75.7	&82.2	&72.5	&-	&-	&-\\
D3S~\cite{D3S}	    &CVPR2020	  &-	&-	&-	&72.8	&76.8	&66.4	&59.7	&67.6	&46.2\\
SiamCAR~\cite{SiamCAR}	&CVPR2020	  &50.7	&60.0	&51.0	&-	&-	&-	&56.9	&67.0 	&41.5\\
SiamBAN~\cite{SiamBAN}	&CVPR2020	&51.4	&59.8	&52.1	&-	&-	&-	&-	&-	&-\\
DiMP~\cite{DiMP}	    &ICCV2019	&56.9	&65.0	&56.7	&74.0	&80.1	&68.7	&61.1	&71.7	&49.2\\
SiamPRN++~\cite{SiamRPNplusplus}&CVPR2019	&49.6	&56.9	&49.1	&73.3	&80.0	&69.4	&51.7	&61.6	&32.5\\
ATOM~\cite{ATOM}	    &CVPR2019	&51.5	&57.6	&50.5	&70.3	&77.1	&64.8	&55.6	&63.4	&40.2\\
ECO~\cite{ECO}	        &ICCV2017	&32.4	&33.8	&30.1	&55.4	&61.8	&49.2	&31.6	&30.9	&11.1\\
MDNet~\cite{MDNet}	    &CVPR2016	 &39.7	&46.0	&37.3	&60.6	&70.5	&56.5	&29.9	&30.3	&9.9\\
SiamFC~\cite{SiameseFC}	&ECCVW2016	&33.6	&42.0	&33.9	&57.1	&66.3	&53.3	&34.8	&35.3	&9.8\\
\hline
\end{tabular}}
\end{center}
\end{table*}

\subsection{Training Loss}
The prediction head receives $H_x \times W_x$ feature vectors, and outputs $H_x \times W_x$ binary 
classification and regression results. 
We select the prediction of feature vectors corresponding to pixels in the ground-truth bounding box 
as positive samples, the rest are negative samples. 
All samples contribute to the classification loss, while only positive samples contribute to the regression loss. 
In order to reduce the imbalance between positive and negative samples, we down-weigh the loss produced by negative 
samples by a factor 16. 
We employ the standard binary cross-entropy loss for classification, which is defined as
\begin{equation}
\begin{split}
\label{equation:BCE_loss}
\mathcal{L}_{cls} = -\sum_j[y_j{\rm log}(p_j)+(1-y_j){\rm log}(1-p_j)], 
\end{split}
\end{equation}
where $y_j$ denotes the ground-truth label of the $j$-th sample, $y_j=1$ denotes foreground, and $p_j$ denotes the probability belong to  
the foreground predicted by the learned model.
For regression, we employ a linear combination of $\ell_1$-norm loss $\mathcal{L}_{1}(.,.)$ 
and the generalized IoU loss $\mathcal{L}_{GIoU}(.,.)$~\cite{GIoU}. 
The regression loss can be formulated as
\begin{equation}
\begin{split}
\label{equation:bbox_loss}
\mathcal{L}_{reg} = \sum_j\mathbbm{1}_{\{y_j=1\}}[\lambda_{G}\mathcal{L}_{GIoU}(b_j,\hat{b})+\lambda_1\mathcal{L}_1(b_j,\hat{b})]
\end{split}, 
\end{equation}
where $y_j=1$ denotes the positive sample, $b_j$ denotes the $j$-th predicted bounding box, and 
$\hat{b}$ denotes the normalized ground-truth bounding box.
$\lambda_{G} =2 $ and $\lambda_{1} = 5$ are the regularization parameters in our experiments. 

\section{Experiments}
\label{sec-exp}

\subsection{Implementation Details}

{\noindent \textbf{Offline Training. }}
We train our model on the training splits of COCO~\cite{COCO}, TrackingNet~\cite{trackingnet}, LaSOT~\cite{LaSOT}, 
and GOT-10k~\cite{GOT10K} datasets. 
For the video datasets (TrackingNet, LaSOT, and GOT-10k), we directly sample the image pairs from one video sequence 
to collect training samples. 
For COCO detection datasets, we apply some transformations on the original image to generate image pairs. 
The common data augmentation (such as translation and brightness jitter) is applied to enlarge the training set.
The sizes of search region patch and template patch are $256 \times 256$ and $128 \times 128$, respectively. 
The backbone parameters are initialized with \ignore{ResNet-50~\cite{ResNet} pre-trained on ImageNet~\cite{ImageNet}}
ImageNet-pretrained~\cite{ImageNet} ResNet-50~\cite{ResNet}, 
other parameters of our model are initialized with Xavier init~\cite{Xavier}. 
We train the model with AdamW~\cite{AdamW}, setting backbone's learning rate to 1e-5,
other parameters' learning rate to 1e-4, 
and weight decay to 1e-4.
%
We train the network on two Nvidia Titan RTX GPUs with the batch size of 38, for a total of 1000 epochs with 1000 iterations per epoch. 
The learning rate decreases by factor 10 after 500 epochs. 

{\noindent \textbf{Online Tracking. }}
In online tracking, the prediction head outputs 1024 boxes with their confidence scores, 
and then the window penalty is adopted for post-processing these scores. Specifically, the Hanning window with the shape of $32 \times 32$ is applied to scores, weighted by a parameter $w$ (chosen as 0.49 in this work). The final score $score_w$ can be defined as 
\begin{equation}
\begin{split}
\label{equation:window_penalty}
score_w = (1 - w) \times score + w \times score_h
\end{split}, 
\end{equation}
where $score$ is the original score of the tracker's output. $score_h$ is the value of the corresponding position on the Hanning window. 
Based on the window penalty, the confidence of feature points far from the target in the previous frames will be punished. 
Finally, we select the box with the highest confidence score as the tracking result.

\subsection{Evaluation on TrackingNet, LaSOT and GOT-10k Datasets}
In this subsection, we compare our TransT method with twelve state-of-the-art trackers 
published in 2020 (SiamR-CNN~\cite{SiamRCNN}, Ocean~\cite{Ocean}, KYS~\cite{KYS}, DCFST~\cite{DCFST}, 
SiamFC++~\cite{SiamFC++}, PrDiMP~\cite{PrDiMP}, CGACD~\cite{CGACD}, SiamAttn~\cite{DSA}, MAML~\cite{MAML}, D3S~\cite{D3S}, 
SiamCAR~\cite{SiamCAR}, and SiamBAN~\cite{SiamBAN}) and six representative trackers presented before (DiMP~\cite{DiMP}, 
SiamPRN++~\cite{SiamRPNplusplus}, ATOM~\cite{ATOM}, ECO~\cite{ECO}, MDNet~\cite{MDNet} and 
SiamFC~\cite{SiameseFC})\footnote{Many trackers have different variants, such as DiMP50 and DiMP18, 
in the original paper. For fair comparison, We simply select the variant with highest performance. 
For example, DiMP means DiMP50 (DiMP with the ResNet50 backbone) in Table~\ref{tab-sota1}.}.
We report the detailed comparison results on the large-scale LaSOT~\cite{LaSOT}, TrackingNet~\cite{trackingnet}, and 
GOT-10k~\cite{GOT10K} datasets in Table~\ref{tab-sota1}. 

\begin{figure}
\centering
    \includegraphics[width=1.0\linewidth]{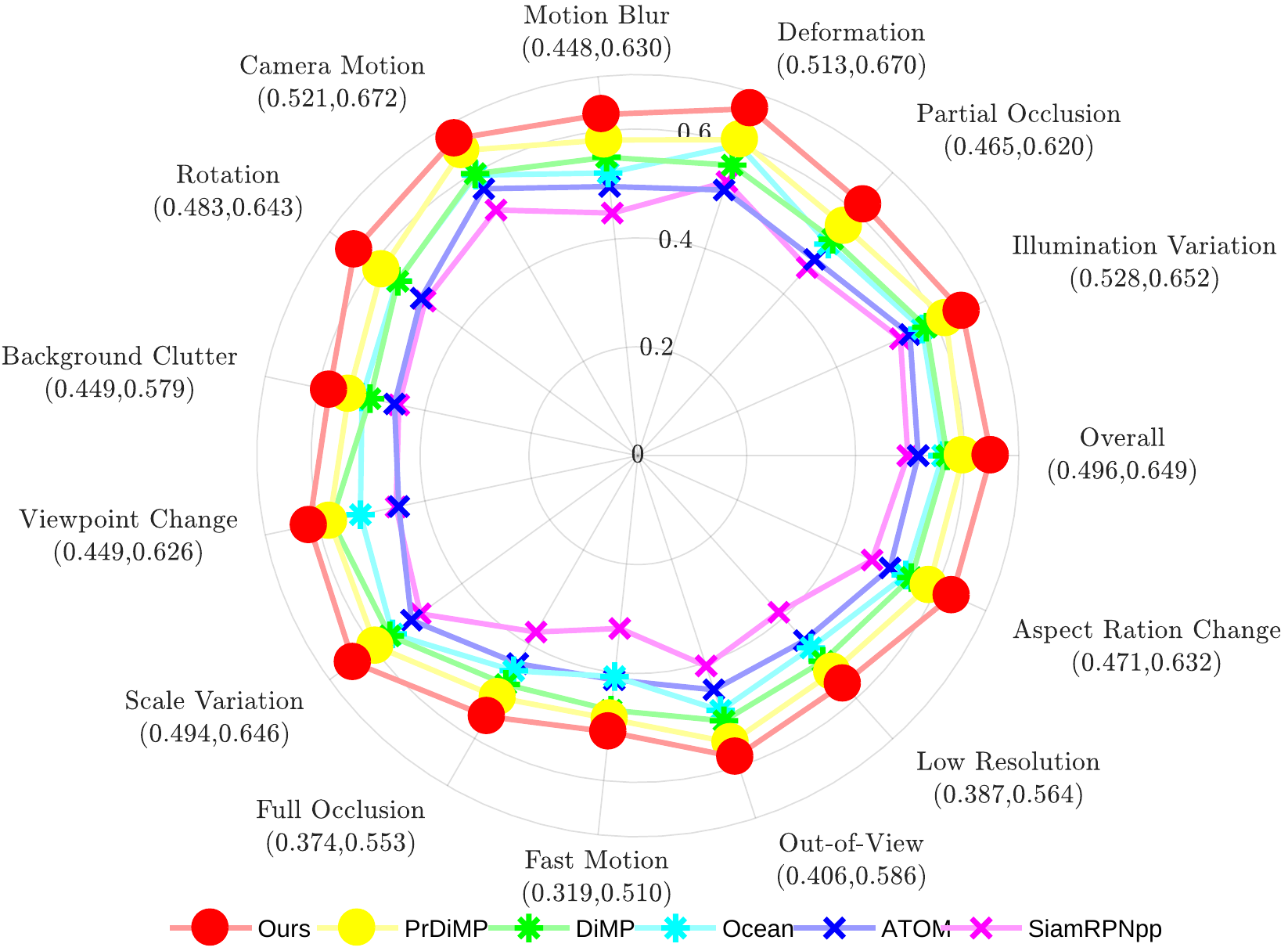}
    \caption{AUC scores of different attributes on the LaSOT dataset.}
    \label{fig:lasotatt}
\end{figure}

{\noindent \textbf{LaSOT. }} LaSOT~\cite{LaSOT} is a recent large-scale dataset with 
high-quality annotations, which contains 1400 challenging videos: 1120 for training and 280 for testing.  
We follow the one-pass evaluation (Success and Precision) to compare different tracking algorithms on the 
LaSOT test set. 
Then, we report the Success (AUC) and Precision (P and $\rm{P}_{Norms}$) scores in Table~\ref{tab-sota1}. 
This table shows that the proposed method obtains the best performance, better than other trackers by a significant margin except SiamR-CNN~\cite{SiamRCNN}, but SiamR-CNN merely runs less than 5\emph{fps} in our machine, while our tracker runs at 50 $fps$.
Figure~\ref{fig:lasotatt} reports an attribute-based evaluation of representative state-of-the-art algorithms, 
illustrating that the TransT performs much better than other competing trackers on all attributes. 

{\noindent \textbf{TrackingNet. }} TrackingNet~\cite{trackingnet} is a large-scale tracking dataset, 
which covers diverse object classes and scenes. Its test set contains 511 sequences publicly available
ground-truth. 
We submit our tracker's outputs to the official online evaluation server, and report the Success (AUC) 
and Precision (P and P$_{Norm}$) results in Table~\ref{tab-sota1}.
our TransT obtains 81.4\%, 86.7\% and 80.3\% in terms of AUC, $\rm{P}_{Norms}$ and P respectively, 
surpassing all previous methods. 

{\noindent \textbf{GOT-10k. }} The GOT-10k~\cite{GOT10K} dataset contains 10k sequences for training and 180 for testing. 
We follow the defined protocol presented in~\cite{GOT10K}, and submit the tracking outputs to the official evaluation 
server.
Then, we report the obtained results (i.e., AO and SR$_{\rm{T}}$) in Table~\ref{tab-sota1}. TransT-GOT denotes training with only the GOT-10k traning set.
TransT and TransT-GOT method achieve the best performance. TransT-GOT method performs 2.2\% higher than SiamR-CNN in the main AO metric.

\subsection{Ablation Study and Analysis}
\label{subsec-as}

\begin{table}[!t]
\caption{Ablation study on TrackingNet, LaSOT, and GOT-10k. The best results 
are shown in the \textbf{\textcolor{red}{red}} font.}
\vspace{-6mm}
\label{tab-ab}
\begin{center}
\resizebox{\linewidth}{!}{
\begin{tabular}{|c|ccc|ccc|ccc|}
\hline
\multirow{2}{*}{Method}  &\multicolumn{3}{|c|}{LaSOT~\cite{LaSOT}}	
&\multicolumn{3}{|c|}{TrackingNet~\cite{trackingnet}}	&\multicolumn{3}{|c|}{GOT-10k~\cite{GOT10K}}\\
\cline{2-10}
&AUC	&P$_{Norm}$	&P	&AUC	&P$_{Norm}$	&P	&AO	&SR$_{0.5}$	&SR$_{0.75}$\\
\hline
TransT &\textcolor{red}{\textbf{64.9}}	&\textcolor{red}{\textbf{73.8}}	&\textcolor{red}{\textbf{69.0}}	
&\textcolor{red}{\textbf{81.4}}	&\textcolor{red}{\textbf{86.7}}	&\textcolor{red}{\textbf{80.3}}	
&\textcolor{red}{\textbf{72.3}}	&\textcolor{red}{\textbf{82.4}}	&\textcolor{red}{\textbf{68.2}}\\
TransT-np       &62.9	&71.5	&66.9	&81.1	&86.4	&80.0	&71.5	&81.5	&67.5\\
TransT(ori)     &62.3	&71.1	&66.2	&81.3	&86.1	&78.9	&70.3	&80.2	&65.8\\
TransT(ori)-np  &60.9	&69.4	&64.8	&80.9	&85.6	&78.4	&68.6	&78.2	&65.1\\
\hline
\end{tabular}
}
\end{center}
\vspace{-7mm}
\end{table}

\begin{table*}[!ht]
\caption{Comparison with correlation on TrackingNet, LaSOT, and GOT-10k. The best results 
are shown in the \textbf{\textcolor{red}{red}} font.}
\vspace{-3mm}
\label{tab-correlation}
\begin{center}
\resizebox{\linewidth}{!}{
\begin{tabular}{|c|c|c|c|ccc|ccc|ccc|}
\hline
\multirow{2}{*}{Method}  &\multirow{2}{*}{ECA} 
&\multirow{2}{*}{CFA} 
&\multirow{2}{*}{Correlation} 
&\multicolumn{3}{|c|}{LaSOT~\cite{LaSOT}}	
&\multicolumn{3}{|c|}{TrackingNet~\cite{trackingnet}}	&\multicolumn{3}{|c|}{GOT-10k~\cite{GOT10K}}\\
\cline{5-13}
& & & &AUC	&P$_{Norm}$	&P	&AUC	&P$_{Norm}$	&P	&AO	&SR$_{0.5}$	&SR$_{0.75}$\\
\hline
TransT &$\surd$ &$\surd$ & &\textcolor{red}{\textbf{64.9}}	&\textcolor{red}{\textbf{73.8}}	&\textcolor{red}{\textbf{69.0}}	
&\textcolor{red}{\textbf{81.4}}	&\textcolor{red}{\textbf{86.7}}	&\textcolor{red}{\textbf{80.3}}	
&\textcolor{red}{\textbf{72.3}}	&\textcolor{red}{\textbf{82.4}}	&\textcolor{red}{\textbf{68.2}}\\
TransT   & &$\surd$ &  &62.9	&71.9	&66.2	&81.1	&86.2	&79.1	&70.6	&81.2	&65.7\\
TransT  &$\surd$ & &$\surd$   &57.7	&65.4	&59.5	&77.5	&82.2	&74.0	&62.8	&72.2	&54.8\\
TransT  & & &$\surd$   &47.7	&48.6	&41.7	&68.8	&71.4	&60.9	&50.9	&58.0	&33.3\\
TransT-np  &$\surd$ &$\surd$ &   &62.9	&71.5	&66.9	&81.1	&86.4	&80.0	&71.5	&81.5	&67.5\\
TransT-np   & &$\surd$ & &61.0	&69.6	&64.5	&80.0	&85.0	&77.9	&68.1	&78.3	&64.0\\
TransT-np  &$\surd$ &  &$\surd$  &57.3	&65.2	&58.8	&76.2	&80.8	&72.8	&61.4	&70.7	&53.7\\
TransT-np  & & &$\surd$  &35.3	&17.9	&20.1	&46.5	&40.3	&27.4	&38.2	&36.8	&7.0\\
\hline
\end{tabular}}
\end{center}
\end{table*}

{\noindent \textbf{Post-processing. }}
In the prior work such as SiamRPN~\cite{SiameseRPN}, SiamRPN++~\cite{SiamRPNplusplus} and Ocean~\cite{Ocean}, 
the final tracking results are selected by post-processing schemes including cosine window penalty, scale change 
penalty and bounding box smoothing. 
However, these post-processing schemes are parameter-sensitive, since three hyparameters that need to 
be adjusted carefully for different test sets. 
To avoid this problem, in this work, we merely adopt the window penalty to conduct post-processing 
using the default parameter for all test sets.

To show the effect of post-processing, we compare the TransT variants with and without the post-processing 
step in Table~\ref{tab-ab}. TransT denotes our tracker and TransT-np is our tracker without post-processing. 
First, from Table~\ref{tab-ab}, we can conclude that our TransT 
without post-processing still achieves state-of-the-art performance, being attributed to the Transformer-like 
fusion method. 
Second, the post-processing step further improves the tracking accuracy, producing the best record among almost all 
metrics on these benchmarks.

{\noindent \textbf{Comparison with the original Transformer. }} 
To show the superiority of our feature fusion network, we design a tracker using the original Transformer. 
Specifically, we replace the feature fusion network in Figure~\ref{fig:TransT} with the original Transformer structure and keep the other components unchanged. 
Because the size of the output of the Transformer is consistent with the size of the decoder input, we input the template feature to the encoder and the search region feature to the decoder. 
The training data and strategy are the same as our TransT in Section~\ref{sec:TransT}. 
The comparison results are shown in Table~\ref{tab-ab}. 
TransT(ori) denotes the tracker with the original Transformer and TransT(ori)-np is the TransT(ori) method 
without post-processing. 
First, the TransT(ori)-np variant achieves an AUC score of 60.9\% on LaSOT, an AUC score of 80.9\% on 
TrackingNet and an AO score of 68.6\% on GOT-10k, which is also better than many state-of-the-art 
algorithms. This indicates that the Transformer structure works better than 
the simple correlation operation in dealing with feature fusion. 
Second, by observing TransT \emph{vs} TransT(ori) and TransT-np \emph{vs} TransT(ori)-np, 
we can conclude that the proposed Transformer performs better than the original Transformer structure, by a large margin. 
Besides, we also see that the post-processing works for both TransT and TransT(ori) methods.

{\noindent \textbf{Comparison with correlation. }} 
Prior Siamese trackers use cross correlation to compute similarity between template and search region.  However, correlation is a linear local comparison, outputting a similarity map. This simple method leads to semantic loss and lacks global information.
Compared with correlation-based methods, first, our attention-based method can establish long-distance feature associations, which effectively aggregates the global information of the template and search region. Second, our method outputs features with rich semantic information, not just a similarity map. 
We conduct experiments to compare CFA with correlation and explore the impact of ECA. To make a fair comparison, for the TransT without CFA, we keep the FFN in CFA unchanged, only remove the cross-attention layers, and replace the last CFA module with depth-wise correlation. The comparison results are shown in Table~\ref{tab-correlation}.
The comparison results show that after replacing CFA with correlation layer, the performance significantly decreases.
Without ECA, the performance of tracker drops. Without both ECA and CFA, the performance further drops, and the impact of post-processing becomes greater.
These results show that without attention modules, the localization ability of the tracker significantly decreases, and it needs to rely more on the prior information in post-processing.


\subsection{Evaluation on Other Datasets} 

\ignore{In Table~\ref{tab-sota-small},} 
We evaluate our tracker on some commonly used small-scale datasets, 
including NFS~\cite{NFS}, OTB2015~\cite{OTB2015}, and UAV123~\cite{UAV}. 
We also collect some state-of-the-art and baseline trackers for comparison. The results are shown in Table~\ref{tab-sota-small}.

{\noindent \textbf{NFS. }}We evaluate the proposed tracker on the 30 fps version of the 
NFS~\cite{NFS} dataset, which contains challenging videos with fast-moving objects. 
The previous best method, PrDiMP, achieves an AUC score of 63.5\%. 
Our method performs better than PrDiMP with a gain of 2.2\%.

{\noindent \textbf{OTB2015. }}OTB2015~\cite{OTB2015} contains 100 
sequences in total and 11 challenge attributes.
Table~\ref{tab-sota-small} shows that our method achieves comparable results with state-of-the-art algorithms (such as PrDiMP and SiamRPN++).

\begin{table}[!t]
\caption{Comparison with state-of-the-art on the OTB100, NFS and UAV123 datasets in terms of overall AUC score. The best two results 
are shown in \textbf{\textcolor{red}{red}} and \textbf{\textcolor{blue}{blue}} fonts.}
\label{tab-sota-small}
\vspace{-6mm}
\begin{center}
\resizebox{\linewidth}{!}{
\begin{tabular}{cc ccc ccc}
\hline
&Ours &PrDiMP~\cite{PrDiMP} &DiMP~\cite{DiMP} &SiamRPN++~\cite{SiamRPNplusplus} &ATOM~\cite{ATOM} &ECO~\cite{ECO} &MDNet~\cite{MDNet}\\
\hline
NFS~\cite{NFS}    &\textcolor{red}{\textbf{65.7}} &\textcolor{blue}{\textbf{63.5}} &62.0 &50.2    &58.4 &46.6 &42.2 \\
OTB~\cite{OTB2015}    &\textcolor{blue}{\textbf{69.4}} &\textcolor{red}{\textbf{69.6}} &68.4 &\textcolor{red}{\textbf{69.6}} &66.9 &69.1 &67.8\\
UAV123~\cite{UAV} &\textcolor{red}{\textbf{69.1}} &\textcolor{blue}{\textbf{68.0}} &65.3 & 61.3   &64.2 &53.2 &52.8\\
\hline
\end{tabular}
}
\end{center}
\vspace{-7mm}
\end{table}

{\noindent \textbf{UAV123. }}UAV123~\cite{UAV} includes $123$ low altitude aerial videos captured from a UAV and adopts success and precision metrics for evaluation.  
As shown in Table~\ref{tab-sota-small}, the proposed method performs the best.

\section{Conclusions}
In this work, we propose a novel, simple, and high-performance tracking framework based on the Transformer-like feature fusion network. 
The proposed network conducts feature fusion solely using the attention mechanism, which includes an ego-context augment module based on 
self-attention and a cross-feature augment module based on cross-attention. 
The attention mechanism establishes long-distance feature associations, making the tracker adaptively focus on useful information and extract abundant semantic information. 
The proposed fusion network could replace correlation to composite the template and search region features, thereby facilitating 
object localization and bounding box regression. 
Numerous experimental results on many benchmarks show that the proposed tracker performs significantly better than the state-of-the-art 
algorithms while running at a real-time speed. 

\vspace{3mm}

{\noindent \textbf{Acknowledgement.}} 
This work was supported in part by the
National Natural Science Foundation of China under Grant
nos. 62022021, 61806037, 61872056, and 61725202, and
in part by the Science and Technology Innovation Foundation of Dalian under Grant no. 2020JJ26GX036.

\clearpage
{\small
\bibliographystyle{ieee_fullname}
\bibliography{egbib}
}

\end{document}